\documentclass[lettersize,journal]{IEEEtran}
\usepackage{amsmath,amsfonts}
\usepackage{algorithmic}
\usepackage{algorithm}
\usepackage{array}
\usepackage[caption=false,font=normalsize,labelfont=sf,textfont=sf]{subfig}
\usepackage{textcomp}
\usepackage{stfloats}
\usepackage{url}
\usepackage{verbatim}
\usepackage{graphicx}
\usepackage{cite}
\usepackage{fontawesome}
\usepackage{booktabs}
\usepackage{multirow}
\usepackage{pifont}

\begin{document}

\title{Event-based Motion Deblurring via Multi-Temporal\\ Granularity Fusion}

 
\author{Xiaopeng Lin\textsuperscript{*} , Hongwei Ren\textsuperscript{*}, Yulong Huang, Zunchang Liu, Yue Zhou, Haotian Fu, Biao Pan\dag, Bojun Cheng\dag
\thanks{* Equal contribution. \dag Corresponding author.}
\thanks{Xiaopeng Lin, Hongwei Ren, Yulong Huang, Zunchang Liu, Yue Zhou, Haotian Fu, and Bojun Cheng are with the MICS Thrust at the Hong Kong University of Science and Technology (Guangzhou).}
\thanks{Biao Pan is at Beihang University.}

}

\markboth{Journal of \LaTeX\ Class Files,~Vol.~14, No.~8, August~2021}%
{Shell \MakeLowercase{\textit{et al.}}: A Sample Article Using IEEEtran.cls for IEEE Journals}


\maketitle

\begin{abstract}
Conventional frame-based cameras inevitably produce blurry effects due to motion occurring during the exposure time. Event camera, a bio-inspired sensor offering continuous visual information could enhance the deblurring performance. Effectively utilizing the high-temporal-resolution event data is crucial for extracting precise motion information and enhancing deblurring performance. However, existing event-based image deblurring methods usually utilize voxel-based event representations, losing the fine-grained temporal details that are mathematically essential for fast motion deblurring. In this paper, we first introduce point cloud-based event representation into the image deblurring task and propose a Multi-Temporal Granularity Network (MTGNet). It combines the spatially dense but temporally coarse-grained voxel-based event representation and the temporally fine-grained but spatially sparse point cloud-based event. To seamlessly integrate such complementary representations, we design a Fine-grained Point Branch. An Aggregation and Mapping Module (AMM) is proposed to align the low-level point-based features with frame-based features and an Adaptive Feature Diffusion Module (AFDM) is designed to manage the resolution discrepancies between event data and image data by enriching the sparse point feature. Extensive subjective and objective evaluations demonstrate that our method outperforms current state-of-the-art approaches on both synthetic and real-world datasets. 
\end{abstract}


\section{Introduction}
\IEEEPARstart{I}{mage} blurring is a common degradation phenomenon due to the existing motions of either moving objects or the camera during the sensor's exposure time  \cite{li2023real, liadaptive}. Traditional image deblurring algorithms rely on mathematical models such as deconvolution to estimate and reverse the blur kernel \cite{fergus2006removing, bahat2017non, kotera2013blind}. It requires prior knowledge about the blur type and is sensitive to light conditions. Recently, deep learning methods have achieved impressive results in image deblurring tasks \cite{cho2021rethinking, tsai2022stripformer, kim2024frequency}. These methods learn effective feature representations from large datasets, showing superior performance in handling complex and unknown blur types with fast processing time. However, frame-based methods which only rely on images captured by conventional cameras show limited performance under challenging light conditions and high-speed complex scenes in real-world scenarios. 

Event-based cameras, such as Dynamic Vision Sensors (DVS) track visual information continuously at each pixel and generate events asynchronously when local brightness changes exceed a certain threshold \cite{brandli2014240, gallego2020event}. This unique mechanism provides high temporal resolution, high dynamic range, and low latency \cite{vitoria2022event}, ideal for image deblurring tasks in dynamic scenes with complex motions or varying light conditions. Moreover, the brightness variations detected in event streams inherently emphasize high-contrast edges, effectively compensating for the loss of fine texture caused by motion blur \cite{zhang2022unifying}. Benefiting from the microsecond accuracy in motion capture, event cameras play an important role in advancing motion deblurring techniques. Recent developments have introduced several innovative event-based deblurring methods \cite{lin2020learning, sun2022event, yang2023event}, enhancing the deblurring effectiveness when combined with the event data.

\begin{figure}[t!]
\centering
\includegraphics[width=0.46\textwidth]{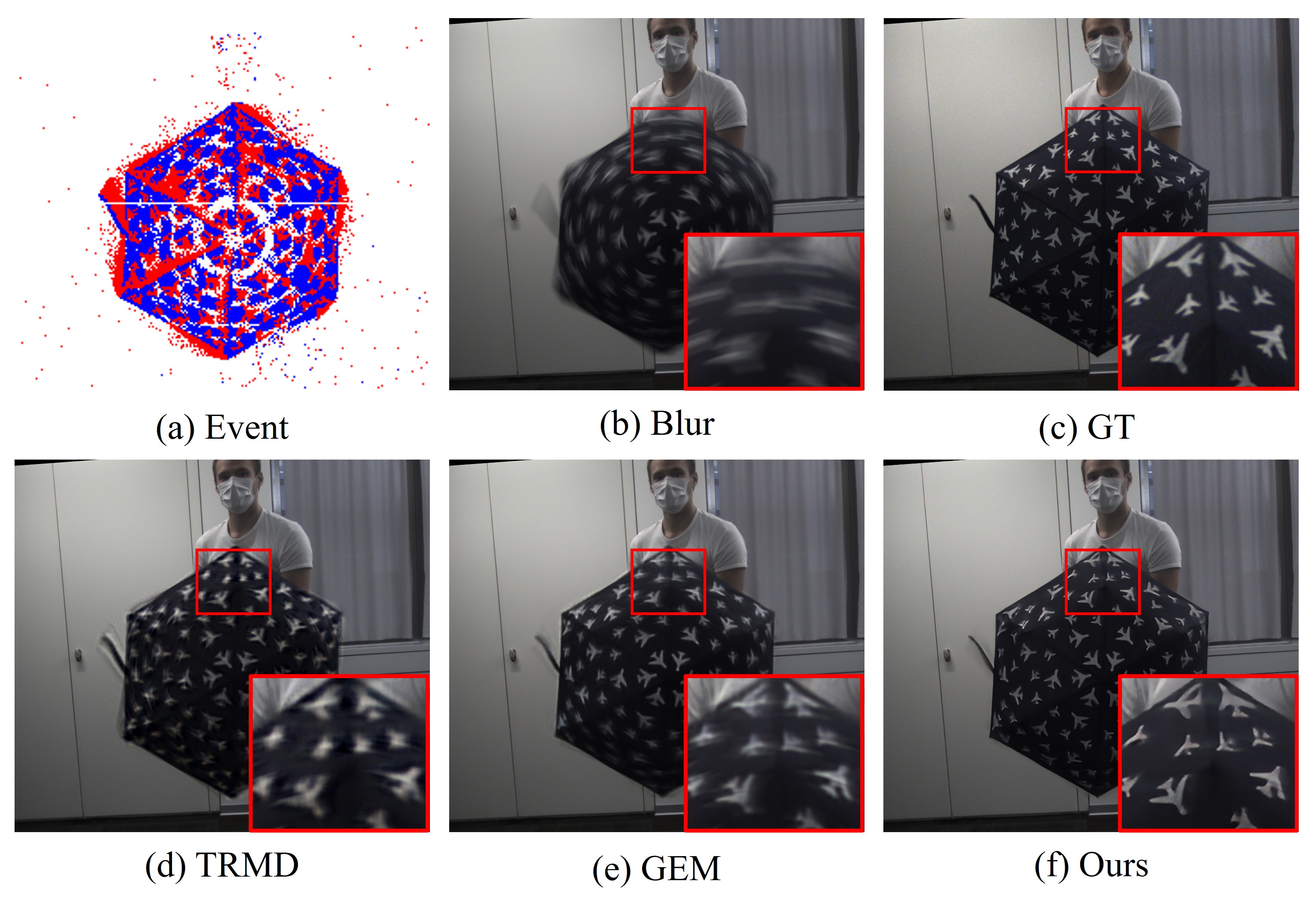}
\caption{Comparison of deblurring results with the state-of-the-art event-based motion deblurring methods, including TRMD, GEM, and our MTGNet.}
\label{fig0}
\end{figure}

The challenge remains in effectively utilizing the asynchronous and sparse event streams. Previous event-based image deblurring works utilize 2D event frames \cite{liu2018adaptive, shang2021bringing, lagorce2016hots} or 3D voxel grids such as Stacking Based on Time (SBT) \cite{wang2019event} and Symmetric Cumulative Event Representation (SCER) \cite{sun2022event} as the event representation for better leveraging advanced neural network backbones. The event frames and voxel-based events accumulate sufficient events in critical pixel coordinates, and the spatial information can be well preserved. However, they compromise the high temporal resolution of raw event data by accumulating counts or polarities \cite{lin2024fapnet}. This reduction in temporal granularity leads to suboptimal deblurring, as it fails to preserve the crucial temporal information for effectively reconstructing deblurred images. To better preserve the fine-grained temporal information of event data, some existing works leverage point cloud as event representation and utilize deep semantic features with event data to achieve impressive results in high-level vision tasks. 
However, point cloud is sparse in the space domain and fine-grained in the time domain. 
It is an advantage for energy-efficient classification tasks but could severely limit the deblurring effectiveness. In fact, aligning the sparse point cloud-based event features with image features presents a notable challenge and there is no point cloud-based method specifically designed for multimodal image deblurring tasks.

In this work, we design a Multi-Temporal Granularity Network (MTGNet) for event-based image deblurring. MTGNet consists of a Coarse Temporal Fusion Branch and a Fine-grained Point Branch. The Fine-grained Point Branch is a complementary branch with rich temporal but sparse spatial information, complimenting the dense spatial and coarse temporal features in the coarse branch. The Fine-grained Point Branch adeptly map the point-like features with frame-like features through the Aggregation and Mapping Module (AMM) and manages resolution discrepancies between event data and image data by introducing the Adaptive Feature Diffusion Module (AFDM). This capability is crucial since event cameras often have lower resolution than frame-based cameras. We also conduct extensive performance comparisons on two datasets with synthetic and real events. Both the subjective and objective evaluations demonstrate our proposed MTGNet outperforms state-of-the-art methods (one exemplary visual comparison is illustrated in Figure \ref{fig0}). The main contributions of our work are as follows:

\begin{itemize}
\item For the first time, point cloud-based event representation is adopted in event-based image deblurring tasks. Combining point cloud and voxel-based representations preserves the temporal details and gives more precise motion information.
\item To enable the deblurring of high-resolution images with relatively low-resolution events, we align the point-based, voxel-based, and frame-based features in space. We also diffuse the point-based feature to enrich the spatially sparse features.
\item Extensive experiments show that equipped with the proposed modules and the multi-temporal granularity event representations, MTGNet achieves SOTA in varying blurry conditions in Ev-REDS, HS-ERGB, and MS-RBD datasets.

\end{itemize}

\section{Related Work}
\subsection{Event Representation}
Event data is asynchronous and sparse, distinct from the synchronous images produced by conventional frame-based cameras. The challenge lies in effectively representing the event stream, which varies depending on the application. Commonly used in fields like action recognition \cite{innocenti2021temporal} and image deblurring \cite{pan2019bringing}, event streams can be concluded into frame-based \cite{shang2021bringing, liu2018adaptive, maqueda2018event}, voxel-based \cite{zhang2022unifying, chen2024motion}, and point cloud-based representations \cite{wang2019space,ren2023ttpoint, lin2024fapnet} to facilitate integration with advanced neural networks, enhancing the applicability across various tasks. Frame-based representation converts events to a 2D frame by counting events or accumulating polarities but it will lose temporal information of event data. Time Surface is also a 2D event map that stores the time value deriving from the timestamp of the last event \cite{sironi2018hats}. The voxel-based representation cuts events into several time bins along the temporal dimension and accumulates the event counts or polarities in each time bin to preserve the coarse temporal information inherent in the events. Stacking Based on Time (SBT) \cite{wang2019event} and Symmetric Cumulative Event Representation (SCER) \cite{sun2022event} are widely used 3D voxel grids event representations in image reconstruction tasks. The point cloud-based representation regards the event data as the point cloud format by treating $t$ to $z$, which can preserve the fine-grained temporal information and better utilize the sparsity and high temporal resolution of event data \cite{ren2024simple}.

\subsection{Point Cloud-based Event Representation Tasks}
Due to the rapid advancement of point cloud networks in 3D vision and the similarity between point cloud and event data, many studies have employed point cloud representations to handle event-based tasks \cite{ma2022rethinking,qian2022pointnext}. PointNet 
\cite{qi2017pointnet} and PointNet++ \cite{qi2017pointnet++} are initially employed for event-based gesture recognition through sliding windows and a downsampling strategy \cite{wang2019space}.
PAT integrates group shuffle attention and Gumbel subset sampling, resulting in enhanced performance in recognition tasks \cite{yang2019modeling}. 
Rasterized EPC is proposed to better represent events in human pose estimation tasks \cite{chen2022efficient}. 
TTPOINT demonstrates outstanding performance in action recognition tasks through its multi-stage architecture integrated with tensor compression \cite{ren2023ttpoint}. Ev2Hands leverages PointNet++ as its backbone for 3D tracking of two fast-moving and interacting hands \cite{millerdurai20243d}. Nevertheless, the fusion of point cloud representation with other modalities has not yet emerged in the event-based vision. The primary challenge lies in the inconsistency between representation methods for point-based features and frame-based features.

\subsection{Image Deblurring}
\subsubsection{Conventional Camera-based Image Deblurring}

In the realm of image deblurring, traditional approaches formulate deblurring as an optimization problem. Conventional CNN-based methods aim to directly learn the mapping from a blurry image to a sharp image through large paired training datasets. MIMO-UNet enhances single image deblurring by employing a multi-input multi-output U-net architecture, enhanced by asymmetric feature fusion for effective multi-scale feature integration \cite{cho2021rethinking}. MSSNet incorporates stage configurations and inter-scale information propagation to enhance both performance and efficiency \cite{wu2024multi}. SSP-Deblur leverages superpixel segmentation prior to guide blind image deblurring, using segmentation entropy as a novel metric to enhance boundary definition and optimize image clarity through a convex energy minimization algorithm \cite{luo2021blind}. LightViD introduces a lightweight video deblurring approach with a blur detector and Spatial Feature Fusion Block, achieving top-tier performance with minimal computational cost \cite{lin2024lightvid}. MSFS-Net proposes a frequency separation module and contrastive learning to refine image details at multiple scales \cite{zhang2023multi}. VDTR introduces a Transformer-based model, utilizing windowed attention and a hierarchical structure to efficiently handle spatial and temporal modeling challenges associated with high-resolution, multi-frame deblurring \cite{cao2022vdtr}.

However, the performance of these methods remains inherently limited when the motion blur occurs outside the scope of the training set, particularly under complex motions or extreme conditions in real-world scenarios.

\subsubsection{Event-based Image Deblurring}

Event cameras provide continuous motion information with low latency, serving as a natural motion clue for enhancing image deblurring. 
Event-based Double Integral (EDI) is proposed to establish a mathematical integral relationship between the blurry image, events, and the sharp reference \cite{pan2019bringing}. 
eSL-Net employs sparse learning to simultaneously denoise and enhance the resolution of images from event cameras, effectively recovering high-quality outputs \cite{wang2020event}. EVDI unifies a framework for event-based motion deblurring and frame interpolation that enhances video quality by utilizing the event camera's low latency to alleviate motion blur and improve frame prediction \cite{zhang2022unifying}. 
DS-Deblur employs a dual-stream framework with adaptive feature fusion and recurrent spatio-temporal transformations \cite{yang2023event}. HybFormer employs hybrid Transformers and a blur-aware detector to effectively utilize both neighboring and interspersed sharp frames for enhanced feature aggregation and frame restoration \cite{ren2023aggregating}. GEM supports variable spatial and temporal scales using a scale-aware network and employs a self-supervised learning strategy to enhance adaptability in diverse real-world environments \cite{zhang2023generalizing}. MotionSNN leverages a spiking neural network and a hybrid feature extraction encoder to enhance event-based image deblurring by effectively integrating high-temporal-resolution event data with traditional image data  \cite{liu2023motion}.

Recent works designing cross-modal attention mechanisms as fusion rules for multimodal integration have achieved impressive results. EFNet introduces a multi-head attention mechanism to fuse information from different modalities and develop a specialized event representation tailored for event-based image deblurring tasks \cite{sun2022event}. EIFNet addresses motion deblurring by decomposing and recomposing modality-specific and modality-shared features, leveraging a dual cross-attention mechanism to separate features and long-range interactions for better fusion \cite{yang2023event}. TRMD utilizes a Two-stage Residual-based Motion Deblurring framework that transforms blurry images into sharp sequences by utilizing motion features from events, incorporating a dual cross-modal fusion module for enhanced residual estimation \cite{chen2024motion}. STCNet employs a differential-modality-based cross-modal calibration strategy to reduce redundancy and a co-attention mechanism for effective spatial fusion, coupled with a frame-event spatial-temporal attention scheme to correct cross-temporal errors \cite{yang2024motion}. 

Although the existing event-based image deblurring methods have yielded notable improvements, there still exist some limitations. Specifically, prior approaches that utilize voxel grids to represent event data have primarily focused on coarse granularity temporal information, neglecting the fine-grained temporal details, thus yielding reduced deblurring performance.


\begin{figure*}[ht]
\centering
\includegraphics[width=0.9\textwidth]{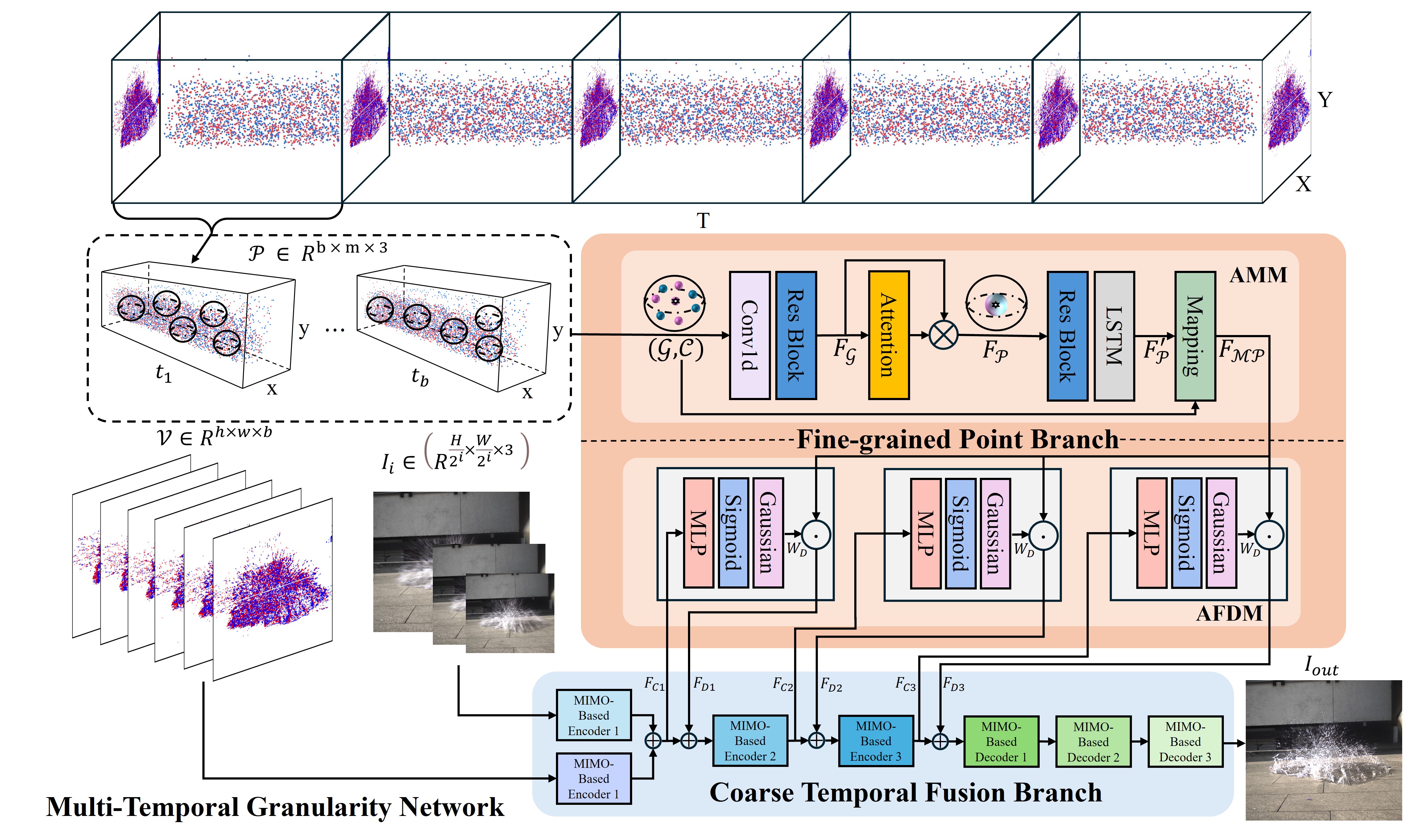}
\caption{The overall framework of MTGNet. The event stream is 
 shaped into the point cloud-based representation $\mathcal{P}$ and the voxel-based representation $\mathcal{V}$. $I_{i}$ are the multi-scale blurry images. MTGNet consists of a Coarse Temporal Fusion Branch and a Fine-grained Point Branch. AMM is the Aggregation and Mapping Module, AFDM is the Adaptive Feature Diffusion Module.}
\label{fig1}
\end{figure*}

\section{Method}
\subsection{Problem Formulation}

Raw events $\mathcal{E}$ captured by the event camera between the generation of blurry images can be described as:
\begin{equation}
        \mathcal{E} = \left\{e_i=(x_i,y_i,t_i,p_i) \mid i=1, \ldots, n\right\},
\end{equation}
where $i$ is the index representing the $i_{\mathrm{\mathrm{th}}}$ event, ($x_i,y_i$) denotes the pixel coordinates that emit an event, $t_i$ represents the timestamp of the event, and $p_i$ is polarity which indicates whether the illumination is brightening or darkening. 

The majority of deblur studies convert the raw event data format to voxel representation $\mathcal{V}$ as the illustrated formula:
\begin{equation}
r_k = t_0 + k\cdot \frac{t_n-t_0}{b},\quad k\in[0,b],
\end{equation}
\begin{equation}
\mathcal{V}=\left\{ v_{x,y}(k) = {\textstyle \sum_{t_i \in (r_k,r_{k+1})} p_i}\mid (t_i,p_i) \in e_i \right\},
\label{eq: voxel}
\end{equation}
where $r$ denotes the rasterized time bin, with a quantity of $b$, $k$ is the index which mean $k_{\mathrm{\mathrm{th}}}$ time bin. The ranges of values for $x$ and $y$ are $[1,w]$ and $[1,h]$, respectively, where $w$ and $h$ represent the resolutions of the event camera.

In contrast, processing raw events using Point Cloud representation simply ignores polarity and treats $t$ as $z$.
\begin{equation}
        \mathcal{P} = \left\{p_i=(x_i,y_i,z_i) \mid i=1, \ldots, n\right\},\quad z=t.
\label{eq: point}
\end{equation}
It can be seen that the two different representations ($\mathcal{V},\mathcal{P}$) have different temporal granularity from Equation \eqref{eq: voxel} and Equation \eqref{eq: point}. Specifically, $\mathcal{P}$ maintains the same microsecond level fine-grained time resolution as $\mathcal{E}$, and the time resolution of $\mathcal{V}$ is only milliseconds level and coarse after rasterization because $b$ is usually set to 6 in previous works.

The blur accumulation process can be modeled by the intensity of sharp images $\mathcal{I}(t)$:
\begin{equation}
    \mathcal{B}=\frac{1}{t_{n}-t_{0}} \int_{t_{0}}^{t_n} \mathcal{I}(t) d t. 
\label{eq： blur process}
\end{equation}
$\mathcal{I}(t)$ can be computed based on the events generated by the current pixel within the interval $[t_0, t]$ as:
\begin{equation}
    \mathcal{I}(t) = \mathcal{I}(t_0)\cdot \text{exp}(\textstyle \sum_{t_j\in(t_0,t)} c_j),
\label{eq： sharp process}
\end{equation}
where $c$ denotes the threshold of intensity change and is a constant determined by hardware.

Substitute Equation \eqref{eq： sharp process} into Equation \eqref{eq： blur process}, we deduce the following formula:
\begin{equation}
\mathcal{I}(t_0) = \frac{\mathcal{B}(t_n-t_0)}{\int_{t_{0}}^{t_n}\text{exp}(\sum_{t_j\in(t_0,t)} c_j)dt}.
\end{equation}
To simplify the formula, we use the function $f(\mathcal{E},t)$ instead of the following module $\text{exp}(\sum_{t_j\in(t_0,t)} c_j)$.

We define the total step for $\int_{t_{0}}^{t_n}$ as $a$ to discrete the deblurring process, so the integral has been rewritten as sum $\sum_{0}^{a}$, the step size $\Delta t$ is $\frac{t_n-t_0}{a}$, the estimated sharp image $\hat{\mathcal{I}}(t_0)$ can be approximately calculated by:
\begin{equation}
\hat{\mathcal{I}}(t_0) = \frac{\mathcal{B}(t_n-t_0)}{\sum_{i=0}^{a} f(\mathcal{E},t_i)\Delta t}.
\end{equation}
The aim of deblurring is to minimize the difference between $\mathcal{I}(t_0)$ and $\hat{\mathcal{I}}(t_0)$, To evaluate the effect of the summation step $\Delta t$ on the performance, we analyze the difference between their reciprocal values. By employing Taylor expansion on the integral, the final difference can be expressed as the following formula:
\begin{equation}
    \frac{1}{\mathcal{I}(t_0)} -\frac{1}{\hat{\mathcal{I}}(t_0)} = \frac{\sum_{i=0}^{a} ( f'(\mathcal{E},t_i)(\Delta t)^2+f''(\mathcal{E},\eta_i)(\Delta t)^3)}{\mathcal{B}(t_n-t_0)},
\label{eqtyler}
\end{equation}
where $\eta_i$ represent a values between $[t_i,t_{i+1}]$, $f'$ and $f''$ are first and second derivatives of function $f$, respectively. It can be concluded that the smaller $\Delta t$, the smaller the difference, and the more similar $\mathcal{I}(t_0)$ and $\hat{\mathcal{I}}(t_0)$ are. In other words, $\Delta t$ represents the temporal granularity of the input data, and the finer $\Delta t$ is, the better the image performs after deblurring.

\subsection{Pre-Processing of the Events}
In this work, we combine the voxel-based representation and Point-based representation event data. The combination more effectively captures the fine-grained temporal dynamics of event data, which can result in better deblurring performance as illustrated in Equation \eqref{eqtyler}.

In the voxel-based representation, the exposure time $T$ of the blurry image is divided equally into $b$ time bins. The event accumulation in each time bin converts the event stream into a 2D tensor with dimensions $(h, w)$, and the voxel-based representation in MTGNet is formulated as defined in Equation \eqref{eq: voxel}.

In this way, $\mathcal{V} \in \mathbb{R}^{h \times w \times b}$ includes the coarse temporal granularity of event data. Moreover, to better reflect real-world conditions, the resolution of event data in our training and test datasets intentionally varies. To mitigate these discrepancies, we apply bicubic interpolation to the voxel-based event representation as a coarse alignment in the spatial dimension.

In the point cloud-based representation, we also first divide the exposure time $T$ into $b$ equal time bins. Instead of stacking events along the temporal dimension, we just downsample fixed number events in each time bin as the uniform format input for the Point Cloud Network. In this study, we randomly select $m$ points from each time interval, and the point cloud-based event of each blurry image is formulated as $\mathcal{P} \in \mathbb{R}^{b \times m \times 3}$. The normalization is then operated in both spatial and temporal dimensions demonstrated as follows:

\begin{equation}
\mathcal{P}_{\mathrm{norm}}^k = (\frac{x_i}{w},\frac{y_i}{h},\frac{t_i-r_k}{r_{k+1}-r_k}),\quad i\in[1,m],k\in[0,b]
\end{equation}
where $\mathcal{P}_{\mathrm{norm}}^k$ is the normalized point cloud-based event data in the $k_{\mathrm{\mathrm{th}}}$ time bin, $x$ and $y$ are the original coordinates, $t_i$ means the original timestamp in the $i_{\mathrm{\mathrm{th}}}$ event. Additionally, to address discrepancies in input resolution, the point cloud is refined by scaling the event coordinates $(x, y)$ according to a specific ratio $\gamma$. This adjustment enhances the precision of calibration across both spatial and temporal dimensions.

\subsection{Network Architecture}
The overall framework of MTGNet is illustrated in Figure \ref{fig1}. The network comprises two branches: the Coarse Temporal Fusion Branch (depicted in the light blue box in Figure \ref{fig1}) and the Fine-grained Point Branch (shown in the light brown box in Figure \ref{fig1}). The Coarse Temporal Fusion Branch is a voxel-based multi-scale UNet, derived from MIMO-UNet \cite{cho2021rethinking}. The Fine-grained Point Branch is designed to leverage the fine-grained temporal information from the event stream. We also focus on addressing the low spatial feature density to effectively integrate both branches.

\subsubsection{Coarse Temporal Fusion Branch}


Extracting and fusing information from event streams and images is essential for effective event-based image deblurring. The Coarse Temporal Fusion Branch employs a three-stage encoder built on the MIMO-UNet architecture. Initially, a Siamese network processes image and event voxel inputs independently, extracting distinct features from each. These extracted features, $F_{\mathcal{I}}$ from images and $F_{\mathcal{V}}$ from event data, are then combined through feature-level addition to produce $F_{\mathcal{C}}$. The initial fusion not only captures and preserves coarse temporal dynamics but also ensures the retention of dense spatial features, laying a solid foundation for accurate motion detection over time. Subsequent to this initial fusion stage, the architecture includes two more stages of the image encoder. These stages process multi-scale images represented as $\mathcal{I} \in (\mathbb{R}^{H \times W \times 3},\mathbb{R}^{\frac{H}{2} \times \frac{W}{2} \times 3},\mathbb{R}^{\frac{H}{4} \times \frac{W}{4} \times 3})$. Progressing from shallow to deep layers, this design effectively handles various types of image blur. The variables $H$ and $W$ denote the height and width, respectively, of the images captured by conventional cameras. This structured approach ensures that each stage contributes optimally to refining the clarity and detail of the deblurred images.

\subsubsection{Fine-grained Point Branch} To better utilize the fine-grained temporal information of event data, we design a Fine-grained Point branch to extract features of point cloud-based event $\mathcal{P}$. The point branch consists of an Aggregation and Mapping Module and an Adaptive Feature Diffusion Module for feature alignment and enhancement.

\textbf{Aggregation and Mapping Module (AMM).} The main challenge of utilizing point cloud-based event data for image deblurring tasks is mapping the point feature with image features, which often come with higher resolution. Previous point cloud-based network preliminary focus on high-level tasks such as action recognition leveraging only the event modality data and does not require precision feature mapping. In this work, we design a coordinate-assisted mapping strategy for better aligning and combining the event point feature with the image and voxel-based event features, as shown in Figure \ref{fig1}. The normalized point cloud $\mathcal{P}_{\mathrm{norm}}$ will be first fed into the Sampling and Grouping module to capture spatio-temporal information. Within each time bin, $\mathcal{P}_{\mathrm{norm}}$ is divided into $M$ groups using Farthest Point Sampling (FPS) and k-nearest neighbors (k-NN) for grouping. The key to the coordinate-assisted mapping strategy is the retention of coordinate position information for the centroid of each group during grouping. It can be formulated as follows:
\begin{equation}
    \{\mathcal{G}_i, \mathcal{C}_i\} = \text{Grouping}(\mathcal{P}_{\mathrm{norm}}, M, K),\quad i\in(1,M)
\end{equation}
where $\mathcal{G}_i$ denotes the $i_{\mathrm{\mathrm{th}}}$ group and $\mathcal{C}_i$ represents the centroid coordinates for the $i_{\mathrm{\mathrm{th}}}$ group, $K$ means the point number in each group and $M$ is the group number per time bin.

We employ a point encoder consisting of convolution layers and res-blocks to refine and elevate the feature set within each group, thereby enriching the point cloud representation. For effective feature mapping, point features within a group are integrated at the centroid using attention-driven weights. This aggregation centralizes the group’s features, effectively augmenting the representation's temporal and spatial characteristics. It can be concluded as follows:

\begin{align}
    F_{\mathcal{G}} = \mathrm{Encoder}&(\mathcal{G}), F_{\mathcal{G}}\in \mathbb{R}^{ M\times K\times D}\\
    F_{\mathcal{P}} = &\sum_{i=1}^{K} \alpha_i \cdot F_{\mathcal{G}}[:,i,:],
\end{align}
where $F_{\mathcal{G}}$ is the point cloud feature sets of the groups, $F_{\mathcal{P}}$ denotes the aggregated feature sets, $\alpha_i$ is the attention weight of the $i_{\mathrm{th}}$ point within the group.

To better integrate information across different time bins, we employ Long Short-Term Memory (LSTM) networks to fuse features along the temporal dimension. Subsequently, the preserved coordinate information from the grouping is utilized to map the aggregated features into the $(x,y)$ plane, enabling effective alignment and fusion with the high-resolution image features and voxel-based event features. It is formulated as follows:

\begin{equation}
\begin{gathered}
    F_{\mathcal{P}}^{\prime} = \text{LSTM}(F_{\mathcal{P}}), \\
    F_{\mathcal{MP}} = \text{Mapping}(F_{\mathcal{P}}^{\prime}, \{\beta_i \cdot C\}),
\end{gathered}
\end{equation}
where $F_{\mathcal{P}}^{\prime}$ is the fusion feature along the temporal dimension, $\beta_i$ is the feature resolution of the $i_{th}$ layer, and $F_{\mathcal{MP}}$ is the mapped feature. 

\textbf{Adaptive Feature Diffusion Module (AFDM).} Due to the excessive sparsity of the mapped point cloud features in the space domain, we design an Adaptive Feature Diffusion Module (AFDM) to enrich the sparse features and manage resolution discrepancies of different vision sensors. The AFDM receives the mapped features $F_{\mathcal{MP}}$ from the AMM and calculates the diffusion range of each pixel by leveraging the initial fusion features $F_{\mathcal{C}}$ from the Coarse Temporal Fusion Branch to enhance the spatial distribution of point features, ensuring a more robust and detailed representation. Specifically, a dynamic diffusion range is first estimated from the blurry image features. The adaptive dilation further determines the extent of feature influence on adjacent points more precisely, which is computed by applying a Multilayer Perceptron (MLP). It is formulated as:
\begin{equation}
\begin{gathered}
     F_{\mathcal{C}}^{\prime} = \frac{1}{H \times W} \sum_{i=1}^{H} \sum_{j=1}^{W} F_{\mathcal{C}}[:, i, j], \\
     \mathbf{D} = \alpha \cdot \sigma(\text{MLP}(F_{\mathcal{C}}^{\prime})),
\end{gathered}
\end{equation}
where $F_{\mathcal{C}}$ denotes the initial fusion features from the Coarse Temporal Fusion Branch, $F_{\mathcal{C}}^{\prime}$ is the features after global average pooling, $\alpha$ is a predefined maximum diffusion range, and it is set to 5 in this study. $\sigma$ is the sigmoid function ensuring the diffusion range will be normalized between 0 and 1. $\mathbf{D}$ is the diffusion range acts as the standard deviation of the Gaussian function.

Gaussian weights are calculated to control the spatial spread of each point feature's influence. The weights are highly correlated with the distance from the centroid point, which ensures that the influence of feature points decreases exponentially with distance, focusing the impact near the point while allowing a tailored spread. It is defined as:
\begin{equation}
    W_{\mathbf{D}} = \exp\left(-\frac{dist^2}{2 \times \mathbf{D}^2}\right),
\end{equation}
where $W_{\mathbf{D}}$ is the Gaussian weight, $dist$ represents the distance from the centroid point to current diffused points.

\begin{figure*}[t]
\centering
\includegraphics[width=0.8\textwidth]{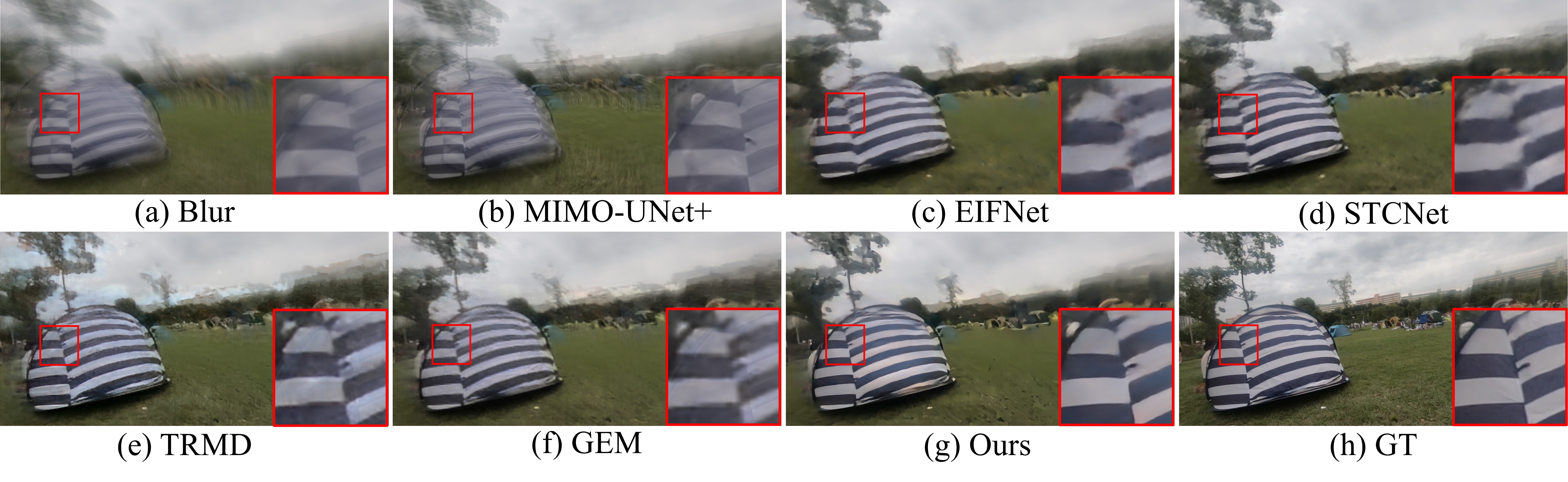}
\caption{Qualitative comparisons under Ev-REDS dataset. Best viewed on a screen and zoomed in \faSearch.}
\label{fig2}
\end{figure*}
\begin{figure*}[t]
\centering
\includegraphics[width=0.8\textwidth]{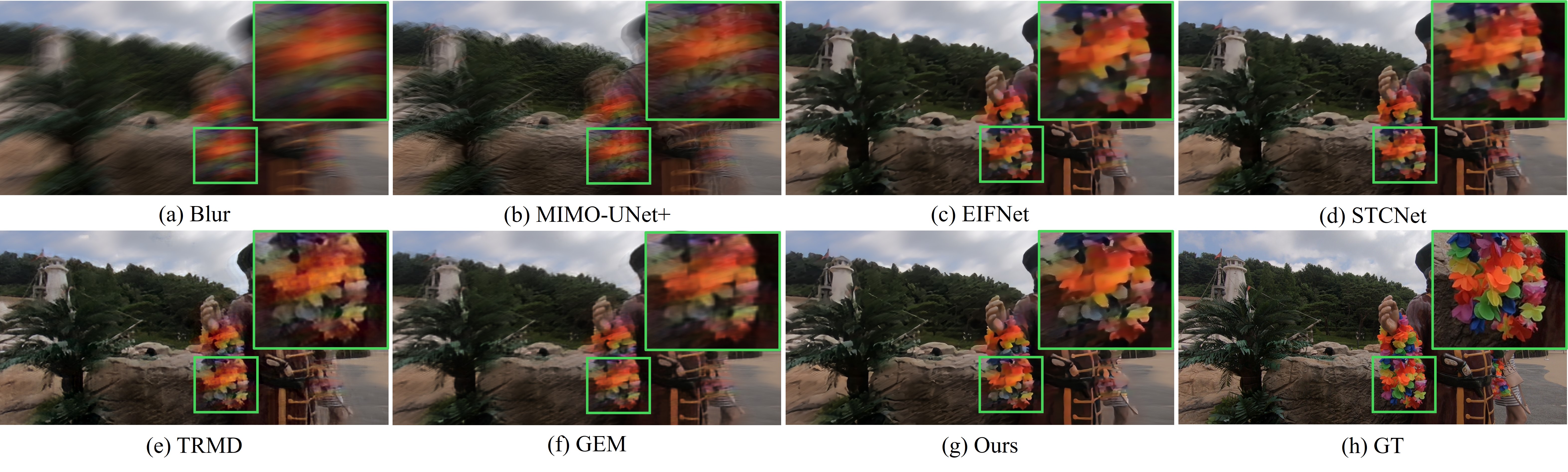}
\caption{Qualitative comparisons under Ev-REDS dataset. Best viewed on a screen and zoomed in \faSearch.}
\label{fig_ev2}
\end{figure*}

The enhanced features via weight diffusion are formulated as follows:
\begin{equation}
    F_{\mathbf{D}} = W_{\mathbf{D}} \odot F_{\mathcal{MP}},
\end{equation}
In MTGNet, diffused point cloud feature $F_{\mathbf{D}}$ and the coarse temporal but spatially dense fusion feature $F_{\mathcal{C}}$ are aggregated in the feature space and processed through a decoder for image reconstruction. The proposed method enhances the deblurring performance by optimally leveraging multiple granularity spatial and temporal information.


\begin{table*}[h]
\centering
\caption{Performance comparison on Ev-REDS and HS-ERGB datasets. The best results are in bold.}
\scalebox{1.0}{
\begin{tabular}{ccccccc}
\toprule
\multirow{2}{*}{Method} & \multicolumn{3}{c}{Ev-REDS} & \multicolumn{3}{c}{HS-ERGB} \\
\cmidrule(r){2-4} \cmidrule(r){5-7}
 & PSNR $\uparrow$ & SSIM $\uparrow$ & LPIPS $\downarrow$ & PSNR $\uparrow$ & SSIM $\uparrow$ & LPIPS $\downarrow$ \\
\midrule
LEVS \cite{jin2018learning}& 18.62 & 0.4612 & \textbackslash & 22.13 & 0.5548 & \textbackslash \\
Motion-ETR \cite{zhang2021exposure}& 18.23 & 0.4292 & \textbackslash & 23.79 & 0.6276 & \textbackslash \\
EVDI \cite{zhang2022unifying}& 23.35 & 0.6368 & \textbackslash & 25.13 & 0.7072 & \textbackslash \\
MIMO-UNet+ \cite{cho2021rethinking}& 20.05 & 0.5416 & 0.1546 & 25.80 & 0.7335 & 0.0849 \\
EIFNet \cite{yang2023event}& 23.88 & 0.6647 & 0.1283 & 26.19 & 0.7328 & 0.0911 \\
GEM \cite{zhang2023generalizing} & 23.95 & 0.6647 & 0.1448 & 26.22 & 0.7292 & 0.0764 \\
TRMD \cite{chen2024motion} & 24.15 & 0.6762 & 0.1330 & 25.81 & 0.7105 & 0.0853 \\
STCNet \cite{yang2024motion} & 24.26 & 0.6887 & 0.1184 & 26.17 & 0.7378 & 0.0841 \\
Ours MTGNet& \textbf{24.52} & \textbf{0.7207} & \textbf{0.0935} & \textbf{27.04} & \textbf{0.7610} & \textbf{0.0669} \\
\bottomrule
\label{table1}
\end{tabular}}
\end{table*}

\subsection{Loss Function}
In this paper, we utilize L1 loss, SSIM loss, and Multi-Scale Frequency Reconstruction (MSFR) loss function to measure the multi-scale pixel-level differences between the estimated image and the ground truth image for the supervised deblurring, formulated as:
\begin{equation}
    L_{\text{total}} = \lambda_1 L_{\text{MAE}} + \lambda_2 L_{\text{SSIM}} + \lambda_3 L_{\text{MSFR}},
\end{equation}
where $\lambda_1=10$, $\lambda_2=1$, and $\lambda_3=0.1$ are the hyper-parameters setting in this work.

\section{Experiment}
\subsection{Experiments Settings}
\textbf{Datasets.} We evaluate the proposed method with Ev-REDS, HS-ERGB, and MS-RBD datasets. (1) We evaluate the deblurring performance on Ev-REDS \cite{zhang2023generalizing} with a focus on global blurry scenarios. Ev-REDS is a widely-used dataset constructed from REDS \cite{nah2019ntire}, which contains synthetic blurry images with corresponding synthetic events generated via VID2E \cite{gehrig2020video}. It includes high-resolution frames at $1280 \times 640$ paired with low-resolution events at $320 \times 160$ for training and testing. (2) We evaluate the deblurring performance on HS-ERGB with a focus on local blurry scenes with more complex moving objects. HS-ERGB dataset \cite{tulyakov2021time} contains sharp videos and real events at the same spatial resolution and the blurry images are generated by using the same strategy as the Ev-REDS. (3) We evaluate the deblurring performance on MS-RBD with a focus on the generalization ability in real-world scenes. MS-RBD \cite{zhang2023generalizing} dataset is created for real-world applicability, which contains high-resolution blurry frames and low-resolution events, capturing the blur effects caused by camera ego-motion and dynamic scenes.


\noindent\textbf{Implementation details.} Our implementation utilizes Pytorch on an NVIDIA RTX 4090 GPU, training on patches of size 512 × 512 with a minibatch size of 1. We employ the ADAM \cite{kingma2014adam} optimizer with an initial learning rate of $1 \times 10^{-4}$, which is scheduled to decrease at the 80th and 100th epochs, over a total of 120 epochs. For data augmentation, we design a density-based random cropping strategy for point cloud-based method. Our evaluation metrics include PSNR, SSIM, and LPIPS.

\noindent\textbf{Density-based Random Cropping Strategy.} Traditional random cropping methods, commonly used for frame-based data, often result in cropped regions containing few or no points when applied to point cloud-based event data. This sparsity significantly limits the effectiveness as a data augmentation strategy for point cloud-based methods. To address this limitation, we introduce a density-based random cropping strategy specifically designed to enhance the suitability of cropped regions for point cloud-based approaches.

The strategy starts by calculating the event density throughout the entire point cloud. Regions where the event density exceeds 80\% are identified to ensure that cropping centers are situated in point-rich areas. From these high-density regions, a central point is selected. And a symmetric expansion is performed around this point to form a $512 \times 512$ patch.

By targeting high-density regions, the proposed strategy ensures that the cropped areas contain sufficient data points, effectively mitigating the issue of sparsity. This approach not only enhances the effectiveness of data augmentation but also optimizes the extracted patches for downstream feature extraction tasks, making it a robust solution for point cloud-based methods.

\noindent\textbf{Network Details.} The detailed parameters of the proposed MTGNet are that: in the \textbf{Pre-Processing of the Events} section, the exposure time $T$ is divided into $b=30$ equal time bins and we randomly select $m=1024$ points in each time interval. If the point counts within the time bin is less than 1024, the events will be repeated until the event number reaches 1024. In this way, the point cloud-based event is formulated as $\mathcal{P} \in \mathbb{R}^{30 \times 1024 \times 3}$, and the last dimension denotes the $(x, y, t)$. The input voxel-based event is formulated as $\mathcal{V} \in \mathbb{R}^{h \times w \times 30}$. In the \textbf{Network Architecture} section, we set the group number $M=1024$ during the Farthest Point Sampling (FPS) and select $K=24$ neighbors for the centroid point of each group. Moreover, the depth of the Resblock layers is set as 20 in the Aggregation and Mapping Module.

\subsection{Comparisons with State-of-the-Art Methods}

We compare our proposed MTGNet to image-only and event-based deblurring methods on Ev-REDS, HS-ERGB, and MS-RBD datasets for a comprehensive evaluation of both local and global blurry scenarios. The comparison methods include image-only method LEVS \cite{jin2018learning}, Motion-ETR \cite{zhang2021exposure}, MIMO-UNet+ \cite{cho2021rethinking}, and event-based methods EVDI \cite{zhang2022unifying}, EIFNet \cite{yang2023event}, GEM \cite{zhang2023generalizing}, TRMD \cite{chen2024motion}, and STCNet \cite{yang2024motion}. For a fair comparison, all methods are trained on the same datasets and under the optimal parameter settings as specified in the respective papers. Our comparison metrics follow the benchmark established by GEM \cite{zhang2023generalizing}, maintaining consistency in metric calculation libraries throughout the study.



$\textbf{Ev-REDS:}$ The detailed evaluation presented in Table \ref{table1} highlights the superior performance of our MTGNet on the Ev-REDS dataset, significantly outperforming comparison methods across key deblurring metrics. MTGNet demonstrates superior performance, achieving a PSNR improvement of 0.57dB over GEM. Our method leads with an SSIM of 0.7207, surpassing STCNet’s SSIM of 0.6887. Most notably, MTGNet also achieves the lowest LPIPS score of 0.0935, indicating a greater perceptual quality compared to STCNet's 0.1184. These metrics collectively underscore MTGNet’s enhanced ability to render clearer and more accurate deblurred images.

The quantitative outcomes are further validated by qualitative assessments shown in Figure \ref{fig2} and Figure \ref{fig_ev2}, where MTGNet demonstrably enhances texture sharpness and edge definition, substantially reducing motion blur relative to other evaluated methods. By leveraging spatially dense voxel-based event features alongside fine-grained temporal data through its Fine-grained Point Branch, MTGNet adeptly addresses complex blurring, underlining its effectiveness and practicality in advancing the quality of deblurred images where conventional approaches might not suffice.
\begin{figure*}[ht]
\centering
\includegraphics[width=0.8\textwidth]{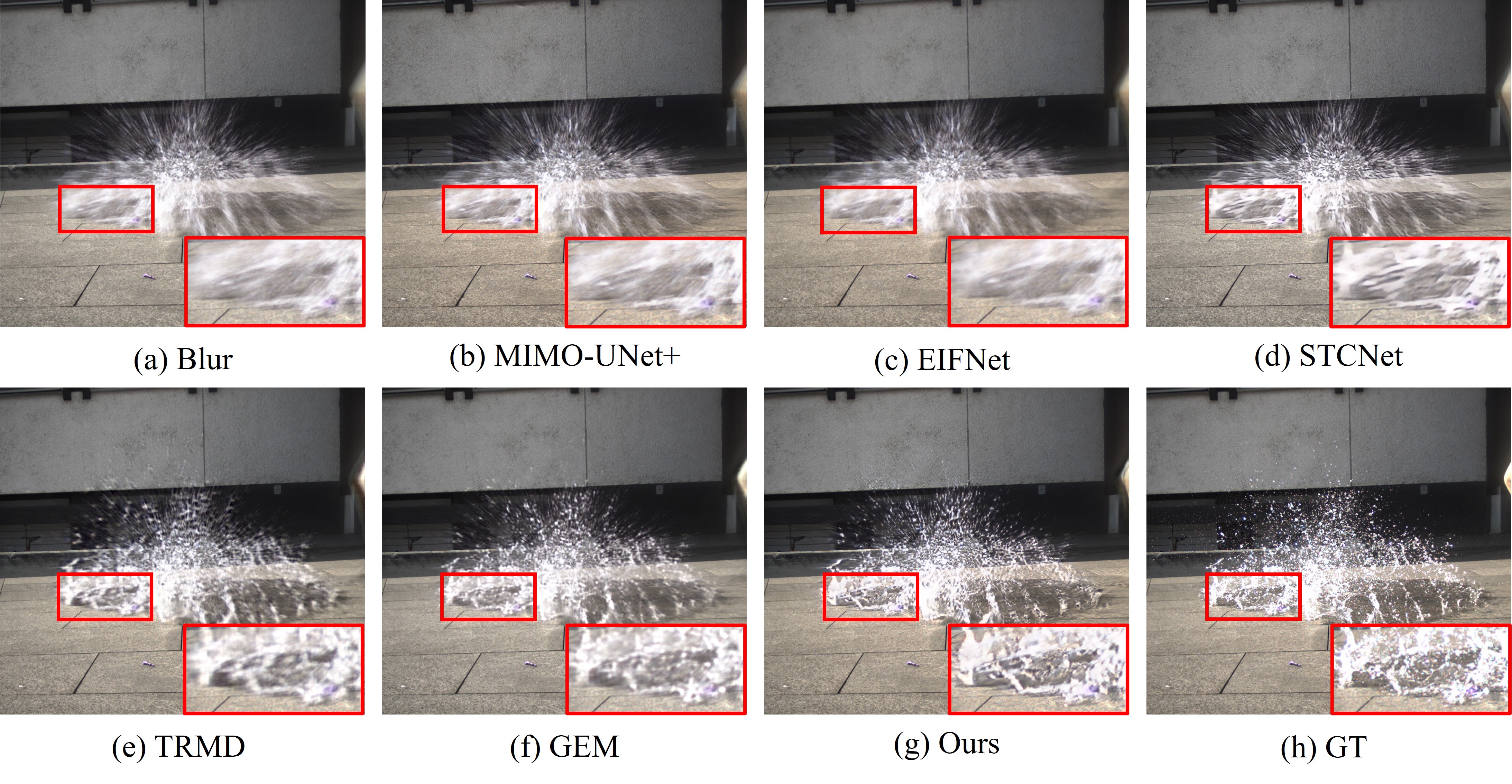}
\caption{Qualitative comparisons under HS-ERGB dataset. Best viewed on a screen and zoomed in.}
\label{fig3}
\end{figure*}


$\textbf{HS-ERGB}:$ In the HS-ERGB evaluation detailed in Table \ref{table1}, our MTGNet distinctly outperforms competing deblurring methods, achieving a leading PSNR of 27.0414, SSIM of 0.7610, and an impressively low LPIPS of 0.0669. These metrics not only underscore a substantial enhancement in image clarity and fidelity but also indicate the superior visual perception capabilities of our method. Specifically, MTGNet excels in preserving intricate details that are often obliterated by the blurring process, as is clearly demonstrated in various comparative analyses. Additionally, MTGNet consistently achieves comparable or superior results against non-open-source methods using the same loss functions and standardized metrics libraries as theirs in the HS-ERGB dataset, confirming its efficacy in image deblurring.


As illustrated in Figure \ref{fig3}, MTGNet achieves better restoration of the complex textures and intricate details within dynamic scenes, exemplified by the splashing water. The zoomed-in sections marked by red boxes in Figure \ref{fig3} showcase MTGNet's deblurring performance in reconstructing the sharp, intricate textures of the water splash. It achieves a level of detail that very closely resembles the ground truth (GT). The fine-grained spatio-temporal details, which are often obscured or lost with other deblurring methods are preserved well in MTGNet. Its ability to handle these challenging deblurring tasks not only demonstrates the effectiveness of its underlying architecture but also highlights its potential for applications requiring high fidelity in dynamic visual environments. This performance is especially crucial in scenarios where clarity and detail retention can significantly impact the utility of the visual output.

\begin{figure*}[ht]
\centering
\includegraphics[width=0.8\textwidth]{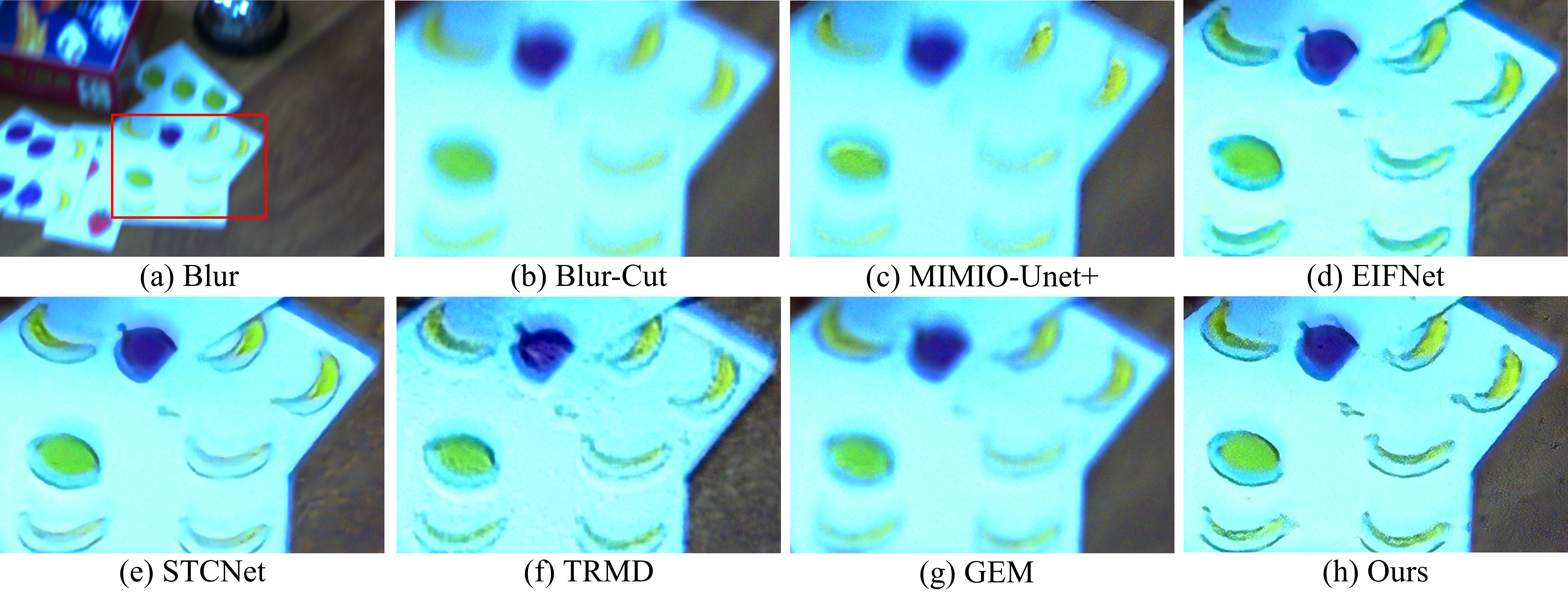}
\caption{Qualitative comparisons under MS-RBD dataset. Best viewed on a screen and zoomed in.}
\label{fig_ms_su}
\end{figure*}

$\textbf{MS-RBD:}$ Figure \ref{fig_ms_su} showcases the qualitative comparison in the MS-RBD dataset which is a real-world scene captured with actual event data and real motion blur. Our method visibly recovers sharper textural details and better preserves the shapes and colors within the scene compared to the other methods. This performance highlights our method's ability to handle complex spatial and color distortions in real-world scenarios. Since the MS-RBD dataset involves real event data and blurry images without corresponding GT images available, all comparison methods are trained on the Ev-REDS dataset and directly applied to the MS-RBD dataset without further adaptation. The subjective evaluation result demonstrates the generalization capabilities of our MTGNet across different datasets and real-world conditions.


The performance of our MTGNet has been thoroughly evaluated across datasets with distinct resolution dynamics, specifically the HS-ERGB dataset with consistent input resolutions, and the Ev-REDS and MS-RBD datasets, where the resolution ratio between RGB and Event data is 4:1. This diverse testing scenario effectively demonstrates MTGNet's exceptional ability to address blurring issues in more realistic situations where sensor resolutions do not align. 

The fusion of multi-time granularity event representations plays a crucial role in achieving these results, particularly in environments with disparate sensor resolutions. Furthermore, the Adaptive Feature Diffusion Module (AFDM) significantly enhances the point cloud-based event features, effectively addressing the challenges posed by high sparsity due to resolution mismatches. The effectiveness of these components in enhancing deblurring accuracy under varied resolution conditions is further explored in the subsequent ablation study, which provides detailed insights into the comparative benefits of integrating fine-grained spatio-temporal information from event data. 

\subsection{Ablation Study}

To evaluate the impact of the multi-temporal granularity and the effectiveness of the key components in the proposed MTGNet, we conduct comprehensive ablation studies on the HS-ERGB datasets, as shown in Table \ref{table2}.

\begin{table}[htbp]
\centering
\caption{Ablation study of the proposed method on HS-ERGB dataset. Image means input blurry image, Voxel stands for voxel-based event representation, Cloud is point cloud-based event representation, and AFDM is the Adaptive Feature Diffusion Module. The best results are in bold.}
\begin{tabular}{cccccc}
\toprule
Image & Voxel & Cloud(AMM) & AFDM & \multicolumn{1}{c}{PSNR / SSIM / LPIPS} \\
\midrule
\ding{51} & & & & 25.80 / 0.734 / 0.0849 \\
\ding{51} & \ding{51} & & & 26.28 / 0.742 / 0.0742 \\
\ding{51} & & \ding{51} & & 25.44 / 0.730 / 0.0984 \\
\ding{51} & & \ding{51} & \ding{51} & 26.74 / 0.753 / 0.0741 \\
\ding{51} & \ding{51} & \ding{51} & & 26.29 / 0.746 / 0.0729 \\
\ding{51} & \ding{51} & \ding{51} & \ding{51} & \textbf{27.04} / \textbf{0.761} / \textbf{0.0669} \\
\bottomrule
\label{table2}
\end{tabular}
\end{table}


\subsubsection{Effectiveness of Multi-Temporal Granularity}

To evaluate the effectiveness of Multi-Temporal Granularity, six ablation experiments are conducted for comparison. 

\textbf{Effectiveness of voxel-based event representation.} Employing only image data without the assistance of event information yields a baseline SSIM of 0.734. The image-only network is the same as the MIMO-UNet \cite{cho2021rethinking}. The incorporation of voxel-based event representation (row 2 in Table \ref{table2}), which encapsulates coarse granularity temporal information but dense spatial details, significantly enhances deblurring performance over the image-only method. 

\begin{figure*}[ht]
\centering
\includegraphics[width=0.8\textwidth]{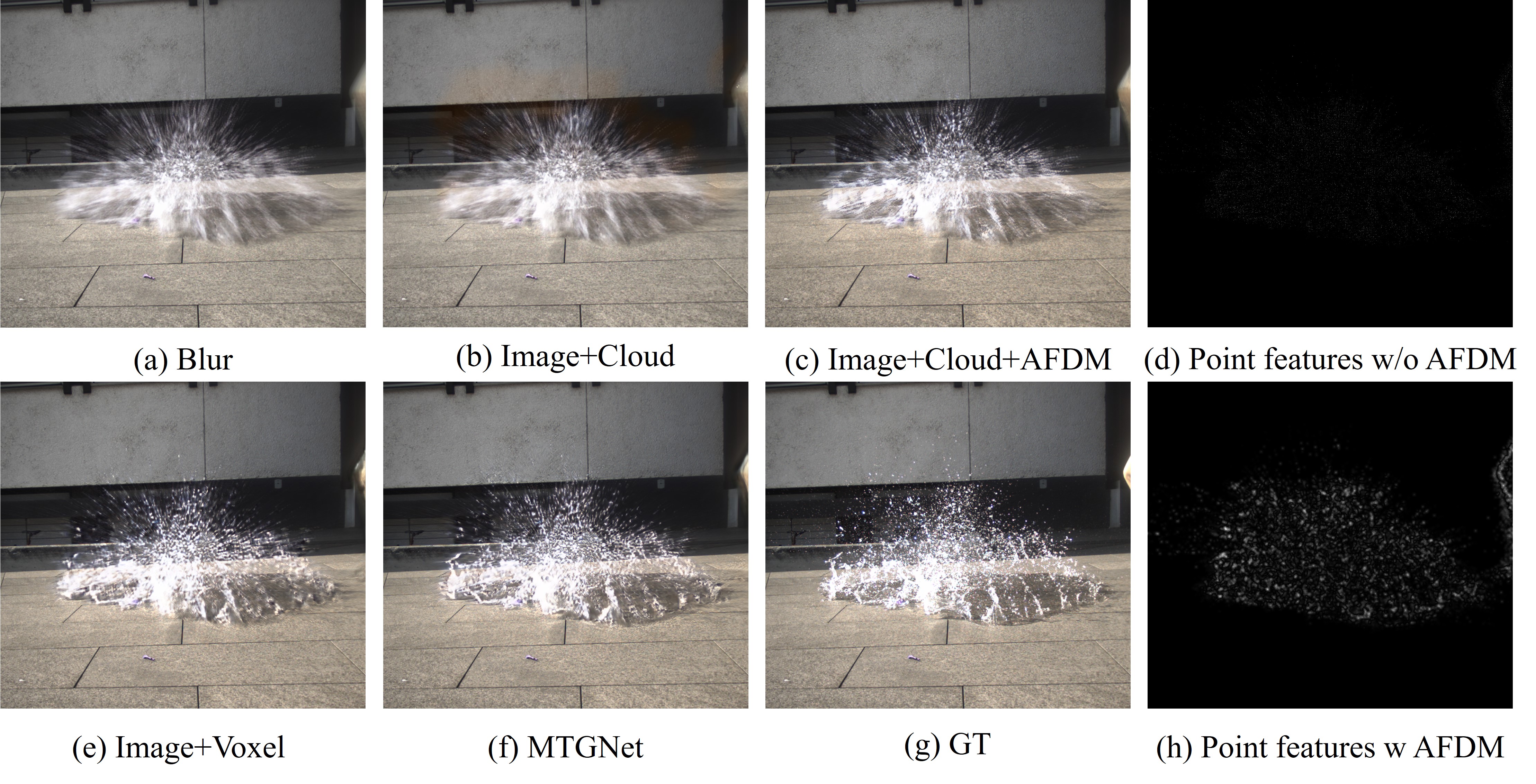}
\caption{Qualitative comparisons of the ablation study. Image means image input, Voxel stands for the voxel-based event representation, Cloud is Point Cloud-based event representation, and AFDM is the Adaptive Feature Diffusion Module. (a) is the blurry image, (b),(c),(e) are the results of the ablation studies with different event representations and modules. (f) is the result of the proposed MTGNet. (d) and (h) are the visualized point cloud-based event features in the feature space.}
\label{fig_ablation}
\end{figure*}

Further analysis between different configurations in Table \ref{table2} reveals how voxel-based event representation contributes under varying conditions. When combined with the temporal fine-grained point cloud-based event, the inclusion of the voxel-based event elevates the SSIM from 0.753 (row 4) to 0.761 (row 6). It indicates a complementary integration between the spatial details provided by the voxel-based representation event and the fine-grained temporal detials from the point cloud-based event representation, enhancing overall image clarity.

\textbf{Effectiveness of point cloud-based event representation.} The comparison of rows 1, 3, and 4 in Table \ref{table2} provides critical insights into the role of point cloud-based event representation in image deblurring tasks. In row 3, when point cloud-based event features that have not undergone feature diffusion, there is a negligible change in performance (PSNR slightly decreases and SSIM remains almost unchanged). This minimal impact can be attributed to the high spatial sparsity of the point cloud after sampling, which leads to a substantial loss of spatial information in the event data. Consequently, the model struggles to capture essential motion clues and effectively perform deblurring tasks. This issue of information loss is further exacerbated when there is a mismatch in spatial resolution between the event data and the image data, complicating the mapping and fusion processes.

Row 4 introduces point cloud data that has been processed through feature diffusion, significantly enhancing model performance (PSNR increases to 26.74 and SSIM to 0.753). This method of feature-level diffusion acts as an interpolation technique that addresses discrepancies in input data resolution. By resolving the sparsity issues of point cloud-based event feature, feature diffusion enables more effective utilization of the point cloud's inherent high temporal resolution. This optimized integration of point cloud data markedly improves the model's ability to discern and clarify motion details, leading to a substantial improvement in deblurring performance, as evidenced by the improved metrics in row 4 compared to row 1 in Table \ref{table2}.

\begin{figure}[ht]
\centering
\includegraphics[width=0.45\textwidth]{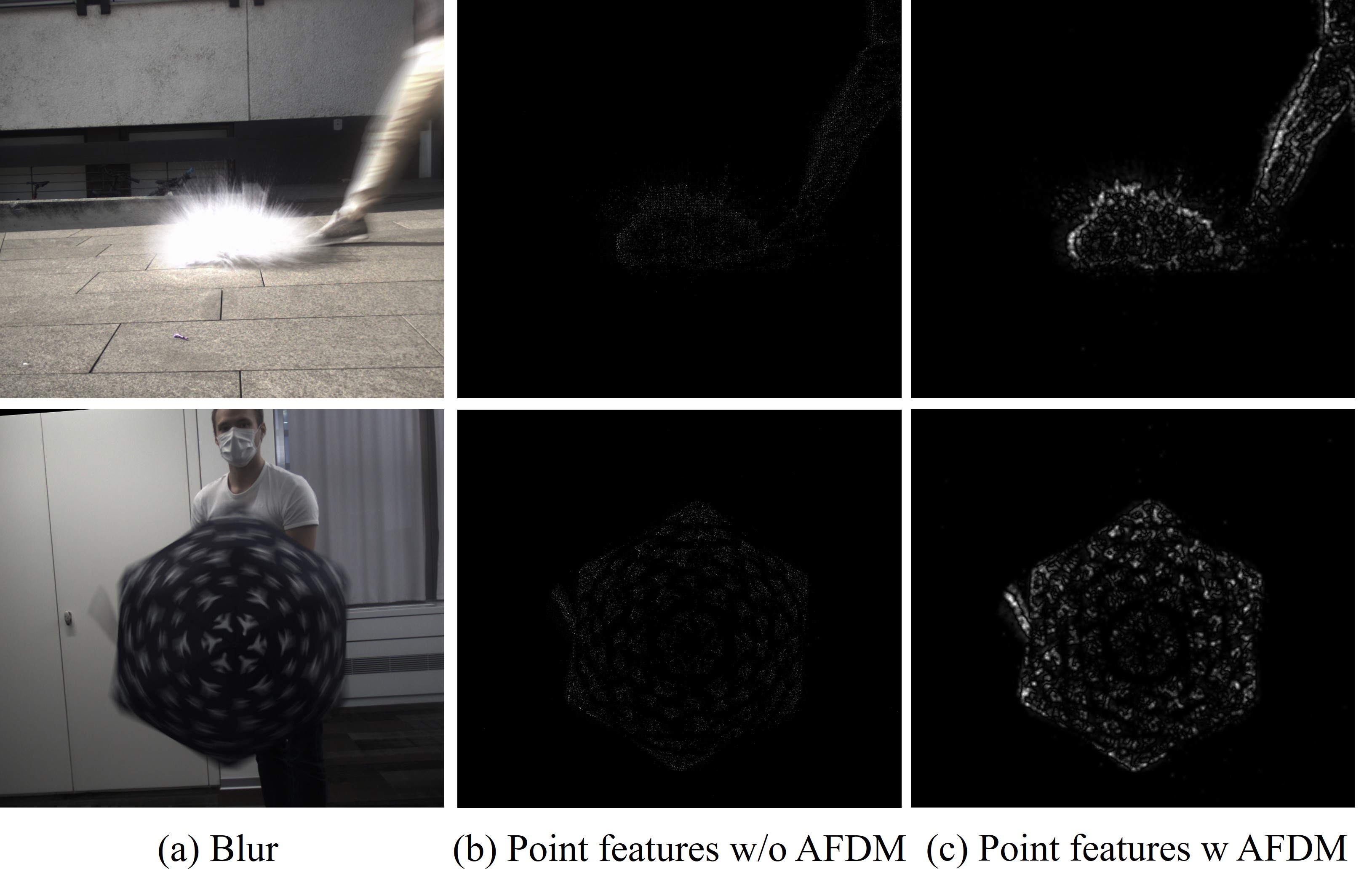}
\caption{Qualitative comparisons of the ablation study with the point feature diffusion. }
\label{fig_ablation_1}
\end{figure}


\subsubsection{Effectiveness of Adaptive Feature Diffusion Module}

The Adaptive Feature Diffusion Module (AFDM) is assessed through a combination of objective metrics and subjective evaluations as illustrated in Table \ref{table2} and Fig. \ref{fig_ablation_1}. The analysis focuses on rows 3, 4, 5, and 6 of Table \ref{table2}, which highlight the significant challenges posed by the sparsity of point cloud features and the effectiveness of AFDM in the deblurring tasks.

Comparing the results of row 3 and row 4 in Table \ref{table2}, before the implementation of feature adaptation through diffusion, the sparse point cloud data fails to preserve crucial spatiotemporal information from event data.
In the real-world scenario, the latest DVS pixel size is ~3-5um, much larger than RGB below 1um so the spatial resolution of event data is much lower than the RGB images. This resolution discrepancy accentuates the information loss in sparse point cloud features, particularly in scenarios where sensor resolutions are misaligned. 

The introduction of AFDM significantly enhances deblurring performance by adaptively diffusing point cloud features to better retain spatiotemporal details. The improvement in performance metrics is evident when comparing the results with and without applying AFDM. For instance in Table \ref{table2}, the transition from row 3 to row 4 shows an increase in PSNR from 25.44 to 26.74 and in SSIM from 0.730 to 0.753, indicating a clearer and more structurally similar image output. Similarly, the enhancement from row 5 to row 6, where PSNR rises from 26.29 to 27.04 and SSIM from 0.746 to 0.761, further confirms the module's effectiveness.

In Figure \ref{fig_ablation_1}, we visualize and compare point cloud features before and after enhancement with the Adaptive Feature Diffusion Module (AFDM). The images distinctly show that post-diffusion, the point cloud features are significantly more robust and densely packed. This enhanced density directly contributes to improved deblurring performance, which can be attributed to two primary factors:
(1)  the data captured by event cameras inherently contains electronic imperfections, such as hot pixels, due to the limitations of sensor circuits. During the sampling process, the noise impacts the acquired points. Without the AFDM, each point cloud feature is treated with equal significance, which means that under sparse conditions, the features are more susceptible to noise interference. The AFDM addresses this by aggregating points within blurry regions, allowing local feature consolidation. Each pixel is then re-weighted according to the diffusion scope, which significantly mitigates the influence of noise. (2)  the feature diffusion process implemented by the AFDM can be considered a form of feature-level interpolation. This adaptation is particularly beneficial in scenarios involving inputs of varying resolutions, enhancing the module's versatility. By employing AFDM, the point cloud features not only retain their high temporal resolution but also achieve spatial feature enhancement. This dual enhancement is crucial for maintaining the temporal accuracy of event data while significantly improving spatial detail, leading to clearer visual outcomes.

\begin{table}[t]
\centering
\caption{Ablation study on HS-ERGB dataset for different ratios of $\lambda_1$:$\lambda_2$:$\lambda_3$ hyperparameters.}
\label{table:hyperparameter_ablation}
\scalebox{1.2}{
\begin{tabular}{cccc}
\hline
\textbf{$\lambda_1$:$\lambda_2$:$\lambda_3$} & 1:1:1 & 1:1:0.1 & \textbf{10:1:0.1} \\ \hline
SSIM & 0.7564 & 0.7595 & \textbf{0.7610} \\
PSNR & 26.9109 & 26.9154 & \textbf{27.0414} \\
LPIPS & 0.0725 & 0.0733 & \textbf{0.0669} \\ \hline
\end{tabular}}
\end{table}

These results demonstrate that AFDM successfully addresses the challenges of data sparsity and resolution mismatch, crucially improving the resolution of event-based data and enabling more effective utilization of the high temporal resolution provided by event cameras. This leads to significantly improved image deblurring, showcasing the practical benefits of feature diffusion in real-world applications.

\subsubsection{Ablation Study of the hyperparameters of loss function}

In our comprehensive ablation study on the hyperparameters of the loss function for MTGNet, we systematically evaluated different parameter ratios to identify the optimal configuration that enhancing deblurring performance. As illustrated in Table \ref{table:hyperparameter_ablation}. We select $\lambda_{1}$:$\lambda_{2}$:$\lambda_{3}$ as 1:1:1, 10:1:0.1, and 1:1:0.1.

As demonstrated in Table \ref{table:hyperparameter_ablation}, the 10:1:0.1 ratio emerged as the most effective, yielding the highest PSNR of 27.0414 and SSIM of 0.7610, along with the lowest LPIPS of 0.0669. This configuration substantially enhances structural similarity and detail recovery, crucial for producing high-quality deblurring. It effectively emphasizes texture and edge preservation while maintaining a balanced contribution from mean absolute error and multi-scale feature reconstruction.

In the study, all ablation experiments have been visually assessed through subjective evaluations showcased in Fig. \ref{fig_ablation}. From the visualizations, it is evident that the Multi-Temporal Granularity event representation significantly enhances deblurring performance. Furthermore, the application of adaptive feature diffusion leads to results that closely approximate the ground truth in visual quality.


 \section{Conclusion}

In this work, we design a novel MTGNet to address motion deblurring by integrating multi-temporal granularity event data and frame images. MTGNet combines spatially dense voxel-based event representation and temporally fine-grained point cloud-based event representation for superior deblurring performance, especially for complex scenes with resolution discrepancy between two sensors. The proposed Fine-grained Point Branch effectively aligns and enriches point features with frame-based features. Comprehensive evaluations across both synthetic and real-world datasets validate that MTGNet is effective in diverse imaging conditions.




\bibliographystyle{IEEEtran}
\bibliography{IEEEabrv,tcsvt}

\begin{thebibliography}{10}
\providecommand{\url}[1]{#1}
\csname url@samestyle\endcsname
\providecommand{\newblock}{\relax}
\providecommand{\bibinfo}[2]{#2}
\providecommand{\BIBentrySTDinterwordspacing}{\spaceskip=0pt\relax}
\providecommand{\BIBentryALTinterwordstretchfactor}{4}
\providecommand{\BIBentryALTinterwordspacing}{\spaceskip=\fontdimen2\font plus
\BIBentryALTinterwordstretchfactor\fontdimen3\font minus \fontdimen4\font\relax}
\providecommand{\BIBforeignlanguage}[2]{{%
\expandafter\ifx\csname l@#1\endcsname\relax
\typeout{** WARNING: IEEEtran.bst: No hyphenation pattern has been}%
\typeout{** loaded for the language `#1'. Using the pattern for}%
\typeout{** the default language instead.}%
\else
\language=\csname l@#1\endcsname
\fi
#2}}
\providecommand{\BIBdecl}{\relax}
\BIBdecl

\bibitem{li2023real}
H.~Li, Z.~Zhang, T.~Jiang, P.~Luo, H.~Feng, and Z.~Xu, ``Real-world deep local motion deblurring,'' in \emph{Proceedings of the AAAI Conference on Artificial Intelligence}, vol.~37, no.~1, 2023, pp. 1314--1322.

\bibitem{liadaptive}
H.~Li, J.~Zhao, S.~Zhou, H.~Feng, C.~Li, and C.~C. Loy, ``Adaptive window pruning for efficient local motion deblurring,'' in \emph{The Twelfth International Conference on Learning Representations}, 2024.

\bibitem{fergus2006removing}
R.~Fergus, B.~Singh, A.~Hertzmann, S.~T. Roweis, and W.~T. Freeman, ``Removing camera shake from a single photograph,'' in \emph{Acm Siggraph 2006 Papers}, 2006, pp. 787--794.

\bibitem{bahat2017non}
Y.~Bahat, N.~Efrat, and M.~Irani, ``Non-uniform blind deblurring by reblurring,'' in \emph{Proceedings of the IEEE international conference on computer vision}, 2017, pp. 3286--3294.

\bibitem{kotera2013blind}
J.~Kotera, F.~{\v{S}}roubek, and P.~Milanfar, ``Blind deconvolution using alternating maximum a posteriori estimation with heavy-tailed priors,'' in \emph{Computer Analysis of Images and Patterns: 15th International Conference, CAIP 2013, York, UK, August 27-29, 2013, Proceedings, Part II 15}.\hskip 1em plus 0.5em minus 0.4em\relax Springer, 2013, pp. 59--66.

\bibitem{cho2021rethinking}
S.-J. Cho, S.-W. Ji, J.-P. Hong, S.-W. Jung, and S.-J. Ko, ``Rethinking coarse-to-fine approach in single image deblurring,'' in \emph{Proceedings of the IEEE/CVF international conference on computer vision}, 2021, pp. 4641--4650.

\bibitem{tsai2022stripformer}
F.-J. Tsai, Y.-T. Peng, Y.-Y. Lin, C.-C. Tsai, and C.-W. Lin, ``Stripformer: Strip transformer for fast image deblurring,'' in \emph{European conference on computer vision}.\hskip 1em plus 0.5em minus 0.4em\relax Springer, 2022, pp. 146--162.

\bibitem{kim2024frequency}
T.~Kim, H.~Cho, and K.-J. Yoon, ``Frequency-aware event-based video deblurring for real-world motion blur,'' in \emph{Proceedings of the IEEE/CVF Conference on Computer Vision and Pattern Recognition}, 2024, pp. 24\,966--24\,976.

\bibitem{brandli2014240}
C.~Brandli, R.~Berner, M.~Yang, S.-C. Liu, and T.~Delbruck, ``A 240$\times$ 180 130 db 3 $\mu$s latency global shutter spatiotemporal vision sensor,'' \emph{IEEE Journal of Solid-State Circuits}, vol.~49, no.~10, pp. 2333--2341, 2014.

\bibitem{gallego2020event}
G.~Gallego, T.~Delbr{\"u}ck, G.~Orchard, C.~Bartolozzi, B.~Taba, A.~Censi, S.~Leutenegger, A.~J. Davison, J.~Conradt, K.~Daniilidis \emph{et~al.}, ``Event-based vision: A survey,'' \emph{IEEE transactions on pattern analysis and machine intelligence}, vol.~44, no.~1, pp. 154--180, 2020.

\bibitem{vitoria2022event}
P.~Vitoria, S.~Georgoulis, S.~Tulyakov, A.~Bochicchio, J.~Erbach, and Y.~Li, ``Event-based image deblurring with dynamic motion awareness,'' in \emph{European Conference on Computer Vision}.\hskip 1em plus 0.5em minus 0.4em\relax Springer, 2022, pp. 95--112.

\bibitem{zhang2022unifying}
X.~Zhang and L.~Yu, ``Unifying motion deblurring and frame interpolation with events,'' in \emph{Proceedings of the IEEE/CVF Conference on Computer Vision and Pattern Recognition}, 2022, pp. 17\,765--17\,774.

\bibitem{lin2020learning}
S.~Lin, J.~Zhang, J.~Pan, Z.~Jiang, D.~Zou, Y.~Wang, J.~Chen, and J.~Ren, ``Learning event-driven video deblurring and interpolation,'' in \emph{Computer Vision--ECCV 2020: 16th European Conference, Glasgow, UK, August 23--28, 2020, Proceedings, Part VIII 16}.\hskip 1em plus 0.5em minus 0.4em\relax Springer, 2020, pp. 695--710.

\bibitem{sun2022event}
L.~Sun, C.~Sakaridis, J.~Liang, Q.~Jiang, K.~Yang, P.~Sun, Y.~Ye, K.~Wang, and L.~V. Gool, ``Event-based fusion for motion deblurring with cross-modal attention,'' in \emph{European conference on computer vision}.\hskip 1em plus 0.5em minus 0.4em\relax Springer, 2022, pp. 412--428.

\bibitem{yang2023event}
W.~Yang, J.~Wu, L.~Li, W.~Dong, and G.~Shi, ``Event-based motion deblurring with modality-aware decomposition and recomposition,'' in \emph{Proceedings of the 31st ACM International Conference on Multimedia}, 2023, pp. 8327--8335.

\bibitem{liu2018adaptive}
M.~Liu and T.~Delbruck, ``Adaptive time-slice block-matching optical flow algorithm for dynamic vision sensors.''\hskip 1em plus 0.5em minus 0.4em\relax BMVC, 2018.

\bibitem{shang2021bringing}
W.~Shang, D.~Ren, D.~Zou, J.~S. Ren, P.~Luo, and W.~Zuo, ``Bringing events into video deblurring with non-consecutively blurry frames,'' in \emph{Proceedings of the IEEE/CVF International Conference on Computer Vision}, 2021, pp. 4531--4540.

\bibitem{lagorce2016hots}
X.~Lagorce, G.~Orchard, F.~Galluppi, B.~E. Shi, and R.~B. Benosman, ``Hots: a hierarchy of event-based time-surfaces for pattern recognition,'' \emph{IEEE transactions on pattern analysis and machine intelligence}, vol.~39, no.~7, pp. 1346--1359, 2016.

\bibitem{wang2019event}
L.~Wang, Y.-S. Ho, K.-J. Yoon \emph{et~al.}, ``Event-based high dynamic range image and very high frame rate video generation using conditional generative adversarial networks,'' in \emph{Proceedings of the IEEE/CVF Conference on Computer Vision and Pattern Recognition}, 2019, pp. 10\,081--10\,090.

\bibitem{lin2024fapnet}
X.~Lin, H.~Ren, and B.~Cheng, ``Fapnet: An effective frequency adaptive point-based eye tracker,'' in \emph{Proceedings of the IEEE/CVF Conference on Computer Vision and Pattern Recognition}, 2024, pp. 5789--5798.

\bibitem{innocenti2021temporal}
S.~U. Innocenti, F.~Becattini, F.~Pernici, and A.~Del~Bimbo, ``Temporal binary representation for event-based action recognition,'' in \emph{2020 25th International Conference on Pattern Recognition (ICPR)}.\hskip 1em plus 0.5em minus 0.4em\relax IEEE, 2021, pp. 10\,426--10\,432.

\bibitem{pan2019bringing}
L.~Pan, C.~Scheerlinck, X.~Yu, R.~Hartley, M.~Liu, and Y.~Dai, ``Bringing a blurry frame alive at high frame-rate with an event camera,'' in \emph{Proceedings of the IEEE/CVF Conference on Computer Vision and Pattern Recognition}, 2019, pp. 6820--6829.

\bibitem{maqueda2018event}
A.~I. Maqueda, A.~Loquercio, G.~Gallego, N.~Garc{\'\i}a, and D.~Scaramuzza, ``Event-based vision meets deep learning on steering prediction for self-driving cars,'' in \emph{Proceedings of the IEEE conference on computer vision and pattern recognition}, 2018, pp. 5419--5427.

\bibitem{chen2024motion}
K.~Chen and L.~Yu, ``Motion deblur by learning residual from events,'' \emph{IEEE Transactions on Multimedia}, 2024.

\bibitem{wang2019space}
Q.~Wang, Y.~Zhang, J.~Yuan, and Y.~Lu, ``Space-time event clouds for gesture recognition: From rgb cameras to event cameras,'' in \emph{2019 IEEE Winter Conference on Applications of Computer Vision (WACV)}.\hskip 1em plus 0.5em minus 0.4em\relax IEEE, 2019, pp. 1826--1835.

\bibitem{ren2023ttpoint}
H.~Ren, Y.~Zhou, H.~Fu, Y.~Huang, R.~Xu, and B.~Cheng, ``Ttpoint: A tensorized point cloud network for lightweight action recognition with event cameras,'' in \emph{Proceedings of the 31st ACM International Conference on Multimedia}, 2023, pp. 8026--8034.

\bibitem{sironi2018hats}
A.~Sironi, M.~Brambilla, N.~Bourdis, X.~Lagorce, and R.~Benosman, ``Hats: Histograms of averaged time surfaces for robust event-based object classification,'' in \emph{Proceedings of the IEEE conference on computer vision and pattern recognition}, 2018, pp. 1731--1740.

\bibitem{ren2024simple}
H.~Ren, J.~Zhu, Y.~Zhou, H.~Fu, Y.~Huang, and B.~Cheng, ``A simple and effective point-based network for event camera 6-dofs pose relocalization,'' in \emph{Proceedings of the IEEE/CVF Conference on Computer Vision and Pattern Recognition}, 2024, pp. 18\,112--18\,121.

\bibitem{ma2022rethinking}
X.~Ma, C.~Qin, H.~You, H.~Ran, and Y.~Fu, ``Rethinking network design and local geometry in point cloud: A simple residual mlp framework,'' \emph{arXiv preprint arXiv:2202.07123}, 2022.

\bibitem{qian2022pointnext}
G.~Qian, Y.~Li, H.~Peng, J.~Mai, H.~Hammoud, M.~Elhoseiny, and B.~Ghanem, ``Pointnext: Revisiting pointnet++ with improved training and scaling strategies,'' \emph{Advances in neural information processing systems}, vol.~35, pp. 23\,192--23\,204, 2022.

\bibitem{qi2017pointnet}
C.~R. Qi, H.~Su, K.~Mo, and L.~J. Guibas, ``Pointnet: Deep learning on point sets for 3d classification and segmentation,'' in \emph{Proceedings of the IEEE conference on computer vision and pattern recognition}, 2017, pp. 652--660.

\bibitem{qi2017pointnet++}
C.~R. Qi, L.~Yi, H.~Su, and L.~J. Guibas, ``Pointnet++: Deep hierarchical feature learning on point sets in a metric space,'' \emph{Advances in neural information processing systems}, vol.~30, 2017.

\bibitem{yang2019modeling}
J.~Yang, Q.~Zhang, B.~Ni, L.~Li, J.~Liu, M.~Zhou, and Q.~Tian, ``Modeling point clouds with self-attention and gumbel subset sampling,'' in \emph{Proceedings of the IEEE/CVF conference on computer vision and pattern recognition}, 2019, pp. 3323--3332.

\bibitem{chen2022efficient}
J.~Chen, H.~Shi, Y.~Ye, K.~Yang, L.~Sun, and K.~Wang, ``Efficient human pose estimation via 3d event point cloud,'' in \emph{2022 International Conference on 3D Vision (3DV)}.\hskip 1em plus 0.5em minus 0.4em\relax IEEE, 2022, pp. 1--10.

\bibitem{millerdurai20243d}
C.~Millerdurai, D.~Luvizon, V.~Rudnev, A.~Jonas, J.~Wang, C.~Theobalt, and V.~Golyanik, ``3d pose estimation of two interacting hands from a monocular event camera,'' in \emph{2024 International Conference on 3D Vision (3DV)}.\hskip 1em plus 0.5em minus 0.4em\relax IEEE, 2024, pp. 291--301.

\bibitem{wu2024multi}
W.~Wu, Y.~Pan, N.~Su, J.~Wang, S.~Wu, Z.~Xu, Y.~Yu, and Y.~Liu, ``Multi-scale network for single image deblurring based on ensemble learning module,'' \emph{Multimedia Tools and Applications}, pp. 1--20, 2024.

\bibitem{luo2021blind}
B.~Luo, Z.~Cheng, L.~Xu, G.~Zhang, and H.~Li, ``Blind image deblurring via superpixel segmentation prior,'' \emph{IEEE Transactions on Circuits and Systems for Video Technology}, vol.~32, no.~3, pp. 1467--1482, 2021.

\bibitem{lin2024lightvid}
L.~Lin, G.~Wei, K.~Liu, W.~Feng, and T.~Zhao, ``Lightvid: Efficient video deblurring with spatial-temporal feature fusion,'' \emph{IEEE Transactions on Circuits and Systems for Video Technology}, 2024.

\bibitem{zhang2023multi}
Y.~Zhang, Q.~Li, M.~Qi, D.~Liu, J.~Kong, and J.~Wang, ``Multi-scale frequency separation network for image deblurring,'' \emph{IEEE Transactions on Circuits and Systems for Video Technology}, vol.~33, no.~10, pp. 5525--5537, 2023.

\bibitem{cao2022vdtr}
M.~Cao, Y.~Fan, Y.~Zhang, J.~Wang, and Y.~Yang, ``Vdtr: Video deblurring with transformer,'' \emph{IEEE Transactions on Circuits and Systems for Video Technology}, vol.~33, no.~1, pp. 160--171, 2022.

\bibitem{wang2020event}
B.~Wang, J.~He, L.~Yu, G.-S. Xia, and W.~Yang, ``Event enhanced high-quality image recovery,'' in \emph{Computer Vision--ECCV 2020: 16th European Conference, Glasgow, UK, August 23--28, 2020, Proceedings, Part XIII 16}.\hskip 1em plus 0.5em minus 0.4em\relax Springer, 2020, pp. 155--171.

\bibitem{ren2023aggregating}
D.~Ren, W.~Shang, Y.~Yang, and W.~Zuo, ``Aggregating long-term sharp features via hybrid transformers for video deblurring,'' \emph{arXiv preprint arXiv:2309.07054}, 2023.

\bibitem{zhang2023generalizing}
X.~Zhang, L.~Yu, W.~Yang, J.~Liu, and G.-S. Xia, ``Generalizing event-based motion deblurring in real-world scenarios,'' in \emph{Proceedings of the IEEE/CVF International Conference on Computer Vision}, 2023, pp. 10\,734--10\,744.

\bibitem{liu2023motion}
Z.~Liu, J.~Wu, G.~Shi, W.~Yang, W.~Dong, and Q.~Zhao, ``Motion-oriented hybrid spiking neural networks for event-based motion deblurring,'' \emph{IEEE Transactions on Circuits and Systems for Video Technology}, 2023.

\bibitem{yang2024motion}
W.~Yang, J.~Wu, J.~Ma, L.~Li, and G.~Shi, ``Motion deblurring via spatial-temporal collaboration of frames and events,'' in \emph{Proceedings of the AAAI Conference on Artificial Intelligence}, vol.~38, no.~7, 2024, pp. 6531--6539.

\bibitem{jin2018learning}
M.~Jin, G.~Meishvili, and P.~Favaro, ``Learning to extract a video sequence from a single motion-blurred image,'' in \emph{Proceedings of the IEEE Conference on Computer Vision and Pattern Recognition}, 2018, pp. 6334--6342.

\bibitem{zhang2021exposure}
Y.~Zhang, C.~Wang, S.~J. Maybank, and D.~Tao, ``Exposure trajectory recovery from motion blur,'' \emph{IEEE Transactions on Pattern Analysis and Machine Intelligence}, vol.~44, no.~11, pp. 7490--7504, 2021.

\bibitem{nah2019ntire}
S.~Nah, S.~Baik, S.~Hong, G.~Moon, S.~Son, R.~Timofte, and K.~Mu~Lee, ``Ntire 2019 challenge on video deblurring and super-resolution: Dataset and study,'' in \emph{Proceedings of the IEEE/CVF conference on computer vision and pattern recognition workshops}, 2019, pp. 0--0.

\bibitem{gehrig2020video}
D.~Gehrig, M.~Gehrig, J.~Hidalgo-Carri{\'o}, and D.~Scaramuzza, ``Video to events: Recycling video datasets for event cameras,'' in \emph{Proceedings of the IEEE/CVF Conference on Computer Vision and Pattern Recognition}, 2020, pp. 3586--3595.

\bibitem{tulyakov2021time}
S.~Tulyakov, D.~Gehrig, S.~Georgoulis, J.~Erbach, M.~Gehrig, Y.~Li, and D.~Scaramuzza, ``Time lens: Event-based video frame interpolation,'' in \emph{Proceedings of the IEEE/CVF conference on computer vision and pattern recognition}, 2021, pp. 16\,155--16\,164.

\bibitem{kingma2014adam}
D.~P. Kingma and J.~Ba, ``Adam: A method for stochastic optimization,'' \emph{arXiv preprint arXiv:1412.6980}, 2014.

\end{thebibliography}

\vfill

\end{document}